\documentclass[letterpaper, 10 pt, conference]{IEEEconf}  %
\pdfoutput=1

\IEEEoverridecommandlockouts   %
\usepackage{graphicx} %

\usepackage[T1]{fontenc}
\usepackage{epsfig} %
\usepackage{etex}
\usepackage{amsmath} %
\usepackage{amssymb}  %
\usepackage{color} %
\usepackage{mathtools}
\usepackage{booktabs}
\usepackage{times}
\usepackage[labelfont=bf, font=footnotesize]{caption}%
\usepackage[labelfont=bf, font=footnotesize]{subcaption}%

\usepackage{enumerate}
\usepackage{balance}
\usepackage{algorithm}
\usepackage{setspace}
\usepackage{varwidth}%
\usepackage{algpseudocode}%
\usepackage{textcomp}
\usepackage{gensymb}
\usepackage{tabulary}
\usepackage[numbers,sort]{natbib}
\usepackage{dblfloatfix}
\usepackage{relsize}
\usepackage{graphbox} %
\usepackage{import}
\usepackage{tabularx}
\usepackage{array}
\usepackage{makecell}
\usepackage{stackengine}
\usepackage{multirow}
\usepackage{dblfloatfix}  %
\usepackage{float}
\usepackage{xargs} %
\usepackage[table, pdftex,dvipsnames]{xcolor}  %
\usepackage[export]{adjustbox}
\usepackage{hyperref}
\usepackage{inconsolata}
\usepackage{url}            %
\let\labelindent\relax %
\usepackage{enumitem}
\usepackage{blindtext}
\usepackage{regexpatch}
\usepackage{tikz} %
\usepackage{bm}

\graphicspath{{images/}}

\hypersetup{
    colorlinks=true,
    filecolor=magenta,      
    urlcolor=magenta,
    citecolor=black,
    pdftitle=ShapeMap 3-D: Efficient shape mapping through dense touch and vision,
    pdfauthor={Sudharshan Suresh, Zilin Si, Joshua G. Mangelson, Wenzhen Yuan, Michael Kaess}
}
\definecolor{query}{HTML}{820D07}
\definecolor{meas}{HTML}{2AA63F}
\definecolor{factor}{HTML}{000287}

\algtext*{EndWhile}%
\algtext*{EndIf}%
\algtext*{EndFor}%

\newcommand*\mystrut[1]{\vrule width0pt height0pt depth#1\relax}
\newcommand{\etal}{et al.~}

\newcommand{\GobbleTiny}[0]{\vspace{-0.1\baselineskip}}
\newcommand{\GobbleSmall}[0]{\vspace{-0.5\baselineskip}}
\newcommand{\GobbleMedium}[0]{\vspace{-1.0\baselineskip}}
\newcommand{\GobbleLarge}[0]{\vspace{-1.5\baselineskip}}

\newcommand{\raisemath}[1]{\mathpalette{\raisem@th{#1}}}
\newcommand{\boldsubheading}[1]{\vspace{0.1in}\noindent\textbf{#1:}}

\newcommand{\cbox}[1]{%
  \begingroup\setlength{\fboxsep}{1pt}%
  \colorbox{lightgray}{\texttt{\hspace*{2pt}\vphantom{y}#1\hspace*{2pt}}}%
  \endgroup
}

\title{\LARGE \bf ShapeMap 3-D: Efficient shape mapping through dense touch and vision}

\author{Sudharshan Suresh$^{*\,1}$, Zilin Si$^{*\,1}$,\\ Joshua G. Mangelson$^2$, Wenzhen Yuan$^1$, and Michael Kaess$^1$%
\thanks{$^{*}$Authors with equal contribution}
\thanks{$^{1}$Sudharshan Suresh, Zilin Si, Wenzhen Yuan, and Michael Kaess are with the Robotics Institute, Carnegie Mellon University {\tt\footnotesize <sudhars1, zsi, wenzheny, kaess>@andrew.cmu.edu}}%
\thanks{$^{2}$Joshua G. Mangelson is with the Electrical and Computer Engineering Department, Brigham Young University {\tt\footnotesize joshua\_mangelson@byu.edu}}%
\thanks{This work was partially supported by the National Science Foundation under award IIS-2008279. We thank Timothy Man for sensor hardware support, and Shubham Kanitkar for help with the robot arm.}
\thanks{Code: \; \; {\tt\url{www.github.com/rpl-cmu/shape-map-3D}}}
\thanks{Dataset: \, {\tt\url{www.github.com/CMURoboTouch/YCB-Sight}}}
}

\makeatletter

\def\maketag@@@#1{\hbox{\m@th\normalfont\normalsize#1}}

\makeatother

\begin{document}

\maketitle
\thispagestyle{empty}
\pagestyle{empty}

\begin{abstract}
Knowledge of 3-D object shape is of great importance to robot manipulation tasks, but may not be readily available in unstructured environments. While vision is often occluded during robot-object interaction, high-resolution tactile sensors can give a dense local perspective of the object. However, tactile sensors have limited sensing area and the shape representation must faithfully approximate non-contact areas. In addition, a key challenge is efficiently incorporating these dense tactile measurements into a 3-D mapping framework. In this work, we propose an incremental shape mapping method using a GelSight tactile sensor and a depth camera. Local shape is recovered from tactile images via a learned model trained in simulation. Through efficient inference on a spatial factor graph informed by a Gaussian process, we build an implicit surface representation of the object. We demonstrate visuo-tactile mapping in both simulated and real-world experiments, to incrementally build 3-D reconstructions of household objects. 
\end{abstract}

\section{Introduction}
\label{sec:intro}
For general-purpose manipulation in unstructured scenes, robots must have accurate understanding of object properties. In particular, knowledge of 3-D shape and its uncertainty enables a breadth of downstream tasks like grasping, dexterous manipulation, and non-prehensile actions. Agents in household or warehouse environments may encounter apriori unknown objects, which they must reconstruct on the fly. 

Vision and depth-based 3-D perception is well-studied \cite{newcombe2011kinectfusion}, but can often fail in the context of manipulation. During some interactions, we only partially observe the scene due to self-occlusion, occlusion from clutter, and fixed viewpoint. Also, visual sensing is degraded by poor illumination, limited range, and ambiguities from transparent or specular objects. 

Studies show humans can optimally fuse touch and vision to reconstruct shape \cite{helbig2007optimal}, reinforcing their complementarity. Vision gives coarse global context, while touch gives precise local information. The development of vision-based tactile sensing \cite{yamaguchi2016combining, yuan2017gelsight, donlon2018gelslim, ward2018tactip, alspach2019soft, lambeta2020digit, padmanabha2020omnitact, wang2021gelsight}, like the GelSight \cite{yuan2017gelsight}, has led to renewed interest in the shape mapping problem. 
Fusing both modalities requires globally integrating tactile signals at the distal end, joint kinematics, and vision. 

Vision-based touch has higher spatial acuity than point-contact or tactile arrays, which lends itself to 3-D reconstruction \cite{wang20183d, bauza2019tactile, smith20203d, smith2021active}. A key challenge is to efficiently incorporating these dense measurements into a 3-D mapping framework. Moreover, the tactile sensor's coverage is limited by its size and durability, and cameras only provide partial visibility of the object. It's desired that a shape representation can faithfully approximate regions lacking measurements. 
\begin{figure}[t]
	\centering
	\includegraphics[width=\columnwidth]{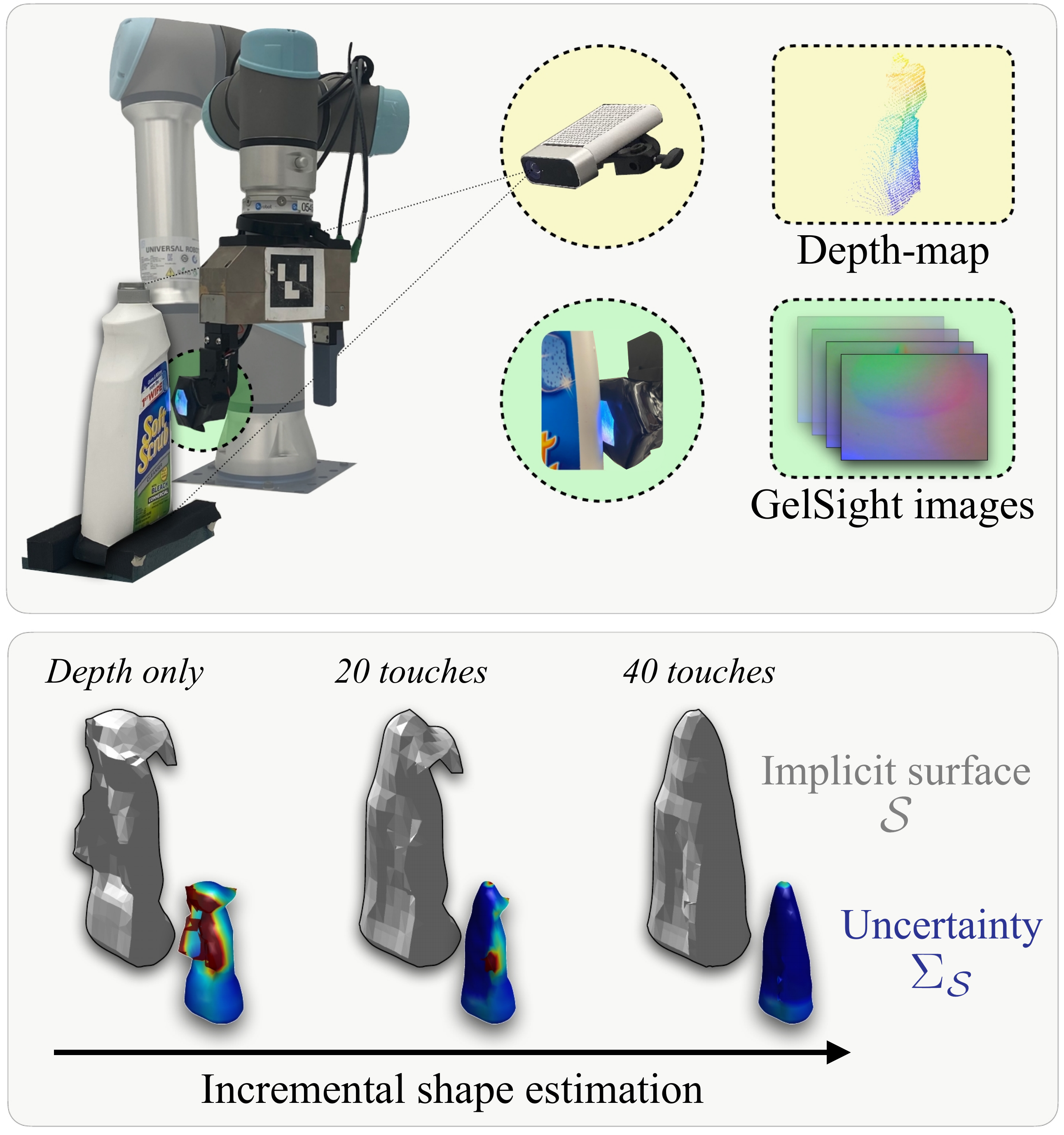}
	\caption{We perform incremental 3-D shape mapping with a vision-based tactile sensor, GelSight, and an overlooking depth-camera. We combine multi-modal sensor measurements into our Gaussian process spatial graph (GP-SG), for efficient incremental mapping. The depth-camera gives us an occluded noisy estimate of 3-D shape, after which we sequentially add tactile measurements as Gaussian potentials into our GP-SG. The tactile measurements are recovered from GelSight images via a learned model trained in simulation. The results demonstrate accurate implicit surface reconstruction and uncertainty prediction for interactive perception tasks.}
	\label{fig:cover} 
	\GobbleLarge
\end{figure}

In this paper, we propose a framework that incrementally reconstructs tabletop 3-D objects from a sequence of tactile images and a noisy depth-map (Figure \ref{fig:cover}). We leverage optical tactile simulation to learn local shape from GelSight-object interactions. We represent 3-D shape as a signed distance function (SDF) sampled from a Gaussian process (GP), and re-formulate shape mapping as probabilistic inference on a spatial graph. We show that visuo-tactile measurements can be incorporated into a factor graph as local Gaussian potentials. This affords efficient access to the implicit surface and SDF uncertainty. We present both simulated and real experiments, generating reconstructions of global shape with trajectories of limited sensor coverage. Specifically, our contributions include:   
\begin{enumerate}
\item[{(1)}] Accurate recovery of local shape from touch learned via tactile simulation of GelSight-object interactions,
\item[{(2)}] Incremental shape mapping through efficient inference in our Gaussian process spatial graph (GP-SG),
\item[{(3)}] Evaluation of visuo-tactile shape mapping on our \texttt{YCBSight-Sim} and \texttt{YCBSight-Real} datasets.
\end{enumerate}

\section{Related work}
\label{sec:related}

\subsection{Tactile sensing and local shape}

For vision-based tactile sensors, photometric stereo \cite{hertzmann2005example} has been widely used to reconstruct local shape \cite{retrographic, microgeometry, yuan2017gelsight}. The approach maps image intensities to gradients via a lookup table, and integrates the gradients to obtain a height-map. However, this method does not consider spatial position in the calibration, and leads to large variance around the boundary of the sensor. Later works learn these gradients directly from tactile images, via either a multilayer perceptron \cite{wang2021gelsight} or pix2pix networks \cite{sodhi2021patchgraph}. Alternatively, end-to-end learning \cite{bauza2019tactile, ambrus2021monocular} from a limited set of real-world tactile interactions can directly provide heightmaps. Tactile simulation \cite{bauza2020tactile, agarwal2021simulation, wang2022tacto, si2022taxim} allows us to scale supervised-learning to a wide range of objects and ground-truth. 

\subsection{Visuo-tactile shape perception}
Global information from vision has complemented low-resolution touch in a multi-modal setting \cite{bjorkman2013enhancing, ilonen2014three, varley2017shape, gandler2020object}. Wang \etal \cite{wang20183d} use monocular shape completion augmented with GelSight readings. However they rely primarily on the visual shape prediction, and tactile sensing serves as a refinement step. Smith \etal \cite{smith20203d, smith2021active} demonstrate a learned perception model on simulated datasets, to predict local mesh deformations via high-resolution touch and filling-in through vision. The context of our work resembles those of \cite{wang20183d} and \cite{smith20203d}, with partial vision and high-dimensional touch. Our contributions differ from these methods as we (i) perform incremental inference on the measurement stream, and (ii) do not rely on data-driven shape priors.

\subsection{Gaussian processes and graphs} 
We wish to faithfully approximate non-contact regions, capture surface uncertainty, and probabilistically handle measurement noise. Gaussian process implicit surfaces (GPIS) \cite{williams2007gaussian} showcase these properties and have found preference in manipulation research---over point-clouds \cite{bauza2019tactile} and other parametric methods \cite{bierbaum2008robust}. The GPIS considers the object's SDF magnitude and gradient as a GP, conditioned on noisy sensor measurements. This has been successfully applied to both passive \cite{dragiev2011gaussian, ottenhaus2016local} and active 3-D reconstruction \cite{bjorkman2013enhancing, jamali2016active, yi2016active, driess2017active} with low-resolution tactile data. We extend these ideas, scaling them to a stream of high-dimensional touch measurements for incremental shape reconstruction. 

The key challenge, especially for GelSight point-clouds, is that GPs scale poorly due to matrix inversion costs. In the SLAM community, common approximations include local GPs \cite{lee2019online, stork2020ensemble} and compact kernels \cite{ranganathan2010online}. These have further been incorporated into factor graphs \cite{dellaert2017factor} for trajectory estimation \cite{yan2017incremental}, target tracking \cite{rosen2014inference}, motion planning \cite{mukadam2018continuous}, elevation modeling \cite{wang2019underwater}, and planar mapping \cite{Suresh21tactile}. Inspired by these, our representation encodes GP potentials as local constraints in a spatial factor graph.

\section{Problem formulation}
\label{sec:prob}

We consider a robot arm with a GelSight tactile sensor interacting with an unknown 3-D object fixed on a tabletop. Given a sequence of images from the GelSight, robot kinematics, and depth-map from a depth-camera, we incrementally estimate the object's shape and signed distance function (SDF) uncertainty. 

\boldsubheading{Object shape}
We represent the object's shape as an implicit surface $\mathbf{\mathcal{S}} \in \mathbb{R}^3$ in the robot's frame, with SDF uncertainty $\mathbf{\Sigma_{\mathcal{S}}}$ (Refer Section \ref{ssec:implicit}).

\boldsubheading{Tactile measurements}
During interaction, upon detecting contact, we record the corresponding tactile image $\mathbf{I_t}$ and sensor pose $\mathbf{p_t}$:
\begin{equation}
\mathbf{z}_t = \Big\{ \mathbf{I}_t \in \mathbb{R}^{\scriptscriptstyle 640\times480\times3}, \ \mathbf{p}_t \in SE\left(3\right)  \Big\}
\label{eq:1}
\end{equation}

\boldsubheading{Depth-map}
We capture a depth-map $\mathbf{D}_0$ of the object from the camera, represented in the robot-frame: $\mathbf{d}_{1 \cdots M} \in \mathbb{R}^{3}$. 

\boldsubheading{Assumptions}
In line with prior efforts, we assume: 
\begin{itemize}[leftmargin=2em]
    \item Calibrated robot-camera extrinsics,
    \item Fixed object pose and known approximate object \\ dimensions,
    \item A passive exploration algorithm for object coverage. 
\end{itemize}

The rest of the paper is as follows: Section \ref{sec:local} presents a GelSight image to height-map model for tactile perception. Section \ref{sec:3d_shape} combines tactile point-clouds with a depth-map in an incremental GP spatial graph. In Section \ref{sec:expts}, we demonstrate our method for simulated and real visuo-tactile experiments. Finally, we sum up our efforts in Section \ref{sec:conc}.

\begin{figure*}[t]
\centering
\begin{minipage}[b]{.46\textwidth}
\includegraphics[width=\textwidth]{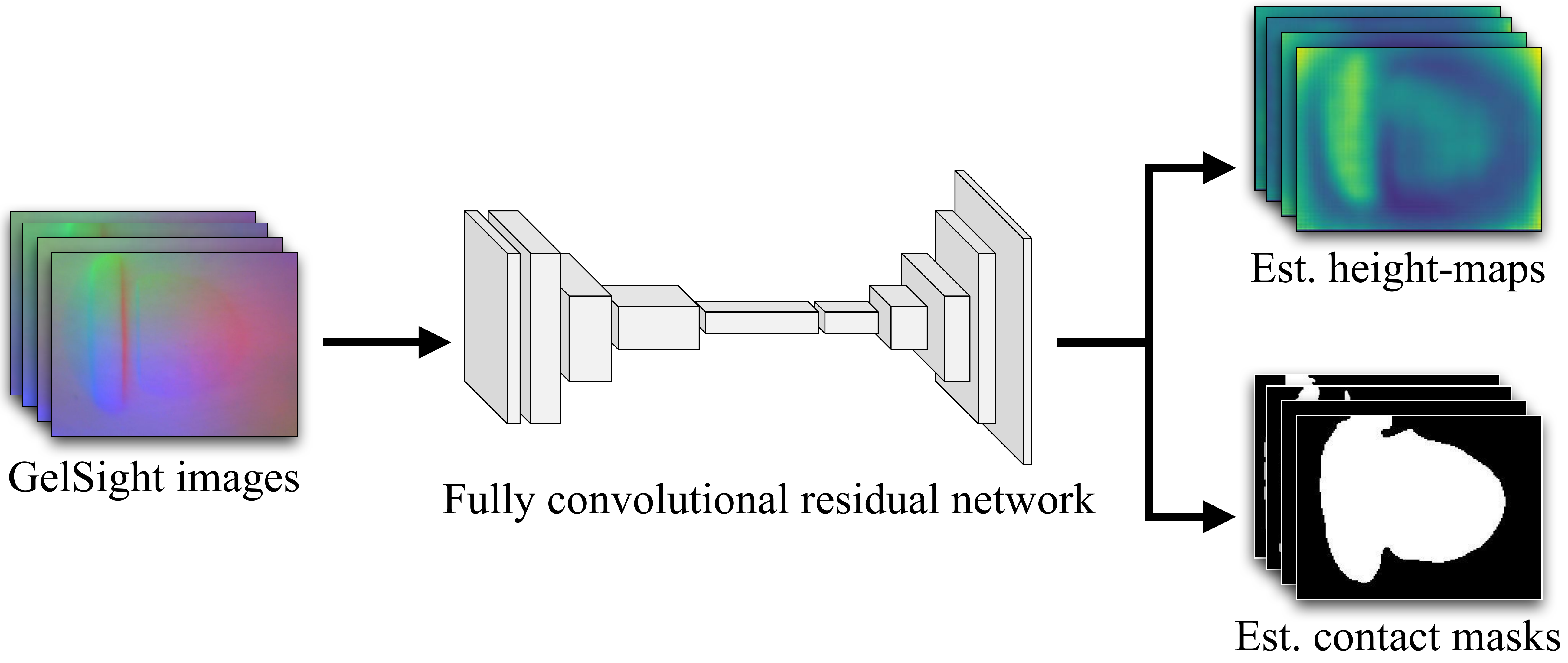}
\caption{Our learned model $\mathbf{\Omega}$ takes in tactile images, and outputs both estimated height-maps and binary contact masks. The residual network is trained on a corpus of GelSight-object interactions in simulation.}	
\label{fig:tactile-flowchart} 
\end{minipage} \quad
\begin{minipage}[b]{.5\textwidth}
\includegraphics[width=\textwidth]{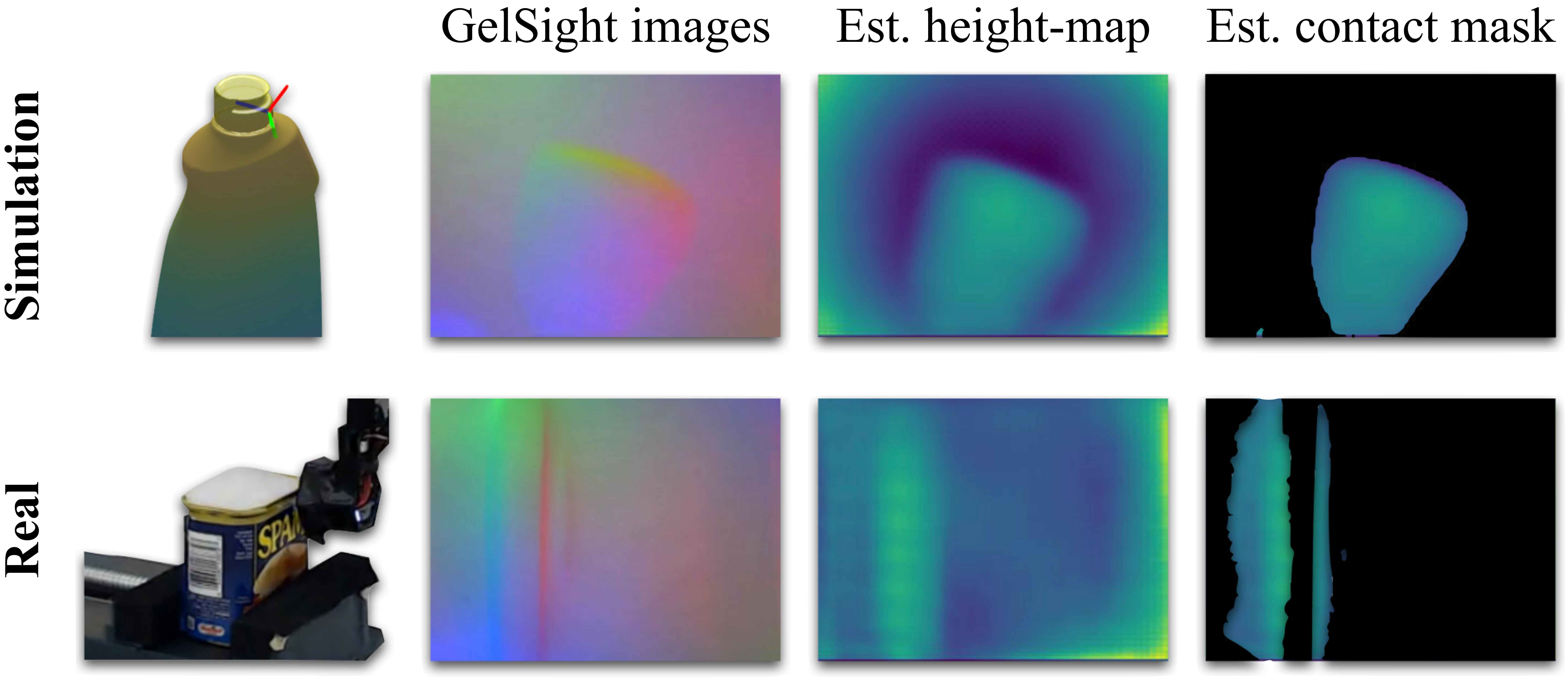}
\caption{Tactile images generated from GelSight interactions in \textbf{[top]} simulated and \textbf{[bottom]} real settings.  Pictured alongside are the height-maps and contact masks output from our learned model $\mathbf{\Omega}$.}
\label{fig:tactile-results} 
\end{minipage}
\GobbleLarge
\end{figure*}
\begin{figure*}[b]
	\GobbleMedium
	\centering
	\includegraphics[width=0.8\textwidth]{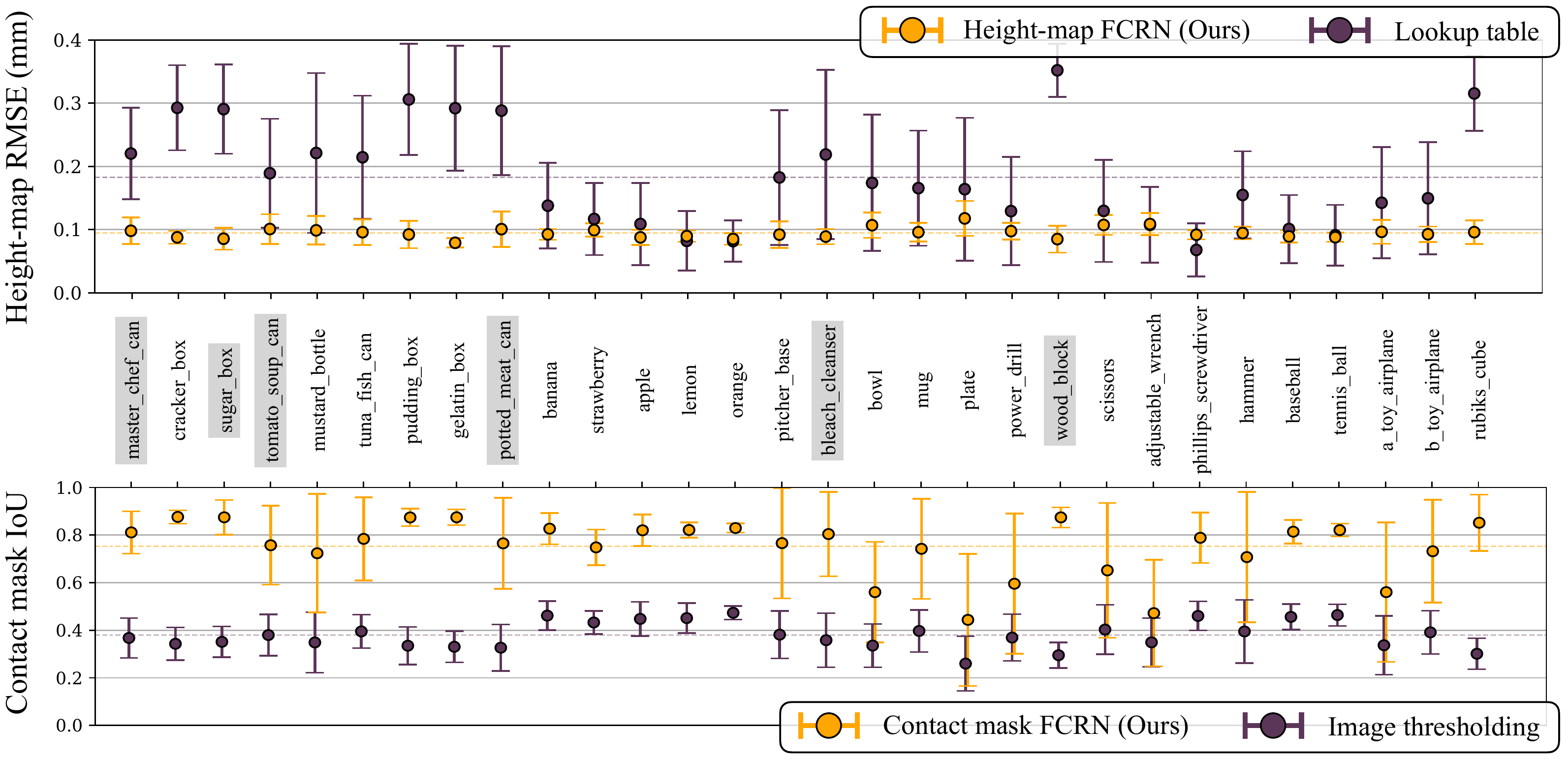}
	\caption{Local shape recovery benchmarked on our \texttt{YCBSight-Sim} dataset (Refer Section \ref{ssec:dataset}). \textbf{[top]} We evaluate our learned model with respect to the baseline lookup table method for height-map estimation. Here we use a pixel-wise root-mean-square error (RMSE)$\downarrow$ metric, and observe consistent, low error for our method when compared with the lookup table. \textbf{[bottom]} We compare our learned contact mask model against intensity-based thresholding on intersection over union (IoU)$\uparrow$ metric. The test data is randomly generated, and \cbox{gray} represents the hold-out objects not encountered in training.}
	\label{fig:tactile-benchmark} 
	\GobbleMedium
\end{figure*}

\section{Local shape from touch}
\label{sec:local}

Vision-based tactile sensors perceive contact geometries as images. The soft, illuminated gelpad deforms elastically on contact and is captured by an embedded camera. We represent local shape recovery as the inverse sensor model: 
\begin{equation}
\mathbf{\Omega} : \mathbf{I}_t \mapsto \mathbf{H}_t, \mathbf{C}_t, \ \text{where} \ \
\parbox{10em}{\raggedright \small $\mathbf{I}_t$: tactile image, \\ 
$\mathbf{H}_t$: recovered height-map, \\
$\mathbf{C}_t$: contact area mask}
\label{eq:2}
\end{equation}
With $\mathbf{H}_t$, $\mathbf{C}_t$, and knowledge of sensor pose $\mathbf{p}_t$ from robot kinematics, we can obtain a tactile point-cloud $\bm{\mathcal{M}}_t$, comprising of 3-D points $\mathbf{m}_{i}^{x}$ and normals $\mathbf{m}_{i}^{y}$:  
\begin{equation}
\bm{\mathcal{M}}_t = \{[\mathbf{m}_{1}^{x}, \mathbf{m}_{1}^{y}], [\mathbf{m}_{2}^{x}, \mathbf{m}_{2}^{y}], \cdots \}
\label{eq:3}
\end{equation}
In this section, we learn $\mathbf{\Omega}$ through simulation, and its output forms the basis for our visuo-tactile mapping in Section \ref{sec:3d_shape}. 

\subsection{Learning  \texorpdfstring{$\boldsymbol{\Omega}$}{Omega} from simulation}
For tactile sensors with soft body deformation, local shape geometry can be learned through supervision. Image-to-depth estimation networks \cite{eigen2014depth,laina2016deeper} can learn accurate heightmaps from GelSight images. This would require a large corpus of tactile images and corresponding ground-truth depths, for which we can leverage tactile simulation. In particular, Si \etal \cite{si2022taxim} calibrate their simulator with real-world tactile images, thus mimicking the same intensity distributions. 

\boldsubheading{Network and training}
We use an implementation \cite{fcrn2018github} of the fully convolutional residual network \cite{laina2016deeper} as our depth estimator, as shown in Figure~\ref{fig:tactile-flowchart}. The network combines ResNet-50 as the encoder and up-sampling blocks as the decoder. Our model takes tactile images as input, and outputs predictions of both height-map $\mathbf{H}_t$ and contact mask $\mathbf{C}_t$. We choose 30 household objects from YCB dataset \cite{calli2017yale}, and hold out 6 objects for testing generalization. For each object, we generate 660 images from randomly sampled sensor poses on their ground-truth mesh models. We split the train-validation-test sets as 550-50-60. 

\boldsubheading{Benchmarks}
We compare $\mathbf{\Omega}$ with the standard lookup table method \cite{yuan2017gelsight}. This maps tactile images to gradients of the local shape, and fast Poisson integration generates their height-maps. The lookup is built via a calibration routine with a $4 \ \text{mm}$ sphere indenter. The contact masks are generated by intensity thresholding of contact vs.\ non-contact frames.

\boldsubheading{Evaluation}
Figure~\ref{fig:tactile-benchmark} compares $\mathbf{\Omega}$ with respect to\ benchmarks on our \texttt{YCBSight-Sim} dataset (refer Section \ref{ssec:dataset}). We compare each estimated height-map and contact mask against the ground-truth. Specifically, we evaluate:
\begin{enumerate}
\item[{(i)}] Pixel-wise RMSE on height-maps, and
\item[{(ii)}] Intersection over union (IoU) on contact masks.
\GobbleTiny
\end{enumerate}
On height-map estimation, we outperform the benchmark with an average RMSE of 0.094 mm across all object classes. The lookup table has larger variance, with an average RMSE of 0.182 mm. Note that the maximum penetration depth of the simulation is 1 mm. On contact mask estimation, we have an average IoU of 0.752, while the handcrafted image thresholding performs worse with 0.379. In addition, the IoU variance appears to be larger for objects with more intricate shapes. Finally, in Figure \ref{fig:tactile-results}, we show generalization of $\mathbf{\Omega}$ to both unseen simulation and real-world tactile interactions.

\section{3-D shape estimation}
\label{sec:3d_shape}
\subsection{Standard Gaussian processes} 
\label{ssec:gp}
A GP is a nonparametric method to learn a continuous function from data, well-suited to model spatial and temporal phenomena \cite{rasmussen2003gaussian}. To estimate shape, a classical GP considers the object's SDF to be a joint Gaussian distribution over noisy measurements of its surface. At any given point in space, the SDF $\varphi$ represents the signed-distance from the surface: $\varphi = 0$ on the surface, $\varphi < 0$ inside, and $\varphi > 0$ outside. The GP meaningfully approximates the global shape, even in regions lacking sensor information. Given a dense tactile measurement $\mathbf{m}_i\in \bm{\mathcal{M}}_t$\footnote{or depth map $\mathbf{d}_{1 \cdots M} \in \mathbf{D}_0$}, we learn a function between positions $\mathbf{m}_{i}^{x}$ and normals $\mathbf{m}_{i}^{y}$:
\begin{equation}
\mathbf{m}_{i}^{x} \mathbf{ \: \mapsto \: } [\varphi = 0, \ \mathbf{m}_{i}^{y}]
\label{eq:4}
\end{equation}

More generally, treating the left and right hand side of Equation \ref{eq:4} as the GP's input-output:
\begin{equation}
\mathbf{X} = \{\mathbf{x}_i \in \, \mathbb{R}^3\}^{1 \cdots N} \mathbf{ \ \mapsto \ }  \mathbf{Y} = \{\mathbf{y}_i  \in  \, \mathbb{R}^4\}^{1 \cdots N} 
\label{eq:5}
\end{equation}
The posterior distribution at a query point $(\mathbf{x}_j^*, \mathbf{y}_j^*)$ for a full GP with $N$ measurements, is given by \cite{rasmussen2003gaussian}:
\footnotesize
\begin{equation}
   \mathbf{y}_j^*
   \sim \mathcal{GP} 
   \big( \underbrace{k_*^T\left(K + \sigma_{\text{n}}^2I \right)^{-1}Y}_{\text{mean}}, \ \
   \underbrace{k_{**} - k_*^T\left(K + \sigma_{\text{n}}^2I \right)^{-1}k_*}_{\text{variance}}
   \big)
   \label{eq:6}
\end{equation}
\normalsize
where $\sigma_{\text{n}}$ is the sensor noise covariance, and $K \in \mathbb{R}^{4N \! \times \! 4N}$, $k_* \in \mathbb{R}^{4N \! \times \! 4}$ and $k_{**} \in \mathbb{R}^{4 \! \times \! 4}$ are the train-train, train-query, and query-query kernels respectively. Each kernel's constituent block $k_{ij} = k(x_i, \ x_j)$ is an $\mathbb{R}^{4 \! \times \! 4}$ kernel basis, in our case a thin-plate function \cite{williams2007gaussian}. This inference is computationally intractable for the large $N$ that accrues from high-dimensional tactile measurements. The update operations involve costly $O(N^3)$ matrix inversions, and per-query costs $O(N^2)$ (Refer Equation \ref{eq:6}). We now present a local approximation that can be updated and queried incrementally, with bounded computational costs.  

\subsection{GP-SG: Gaussian process spatial graph} 
\label{ssec:gpsg}
We represent the scene as a spatial factor graph \cite{dellaert2017factor}, comprising of nodes we optimize for and factors that constrain them. These query nodes $\mathbf{Y}^*$ are at their respective  spatial positions $\mathbf{X}^*$, distributed in an $S^3$ volume. Our optimization goal is to recover the posterior $\hat{\mathbf{Y}}^*$, which represents the SDF of the volume and its underlying uncertainty. 

Implementing the full GP (Equation \ref{eq:6}) in the graph is costly, as each measurement $(\mathbf{x}_i, \mathbf{y}_i)$ constrains all query nodes $\mathbf{Y}^*$. Motivated by prior work in spatial partitioning \cite{lee2019online, stork2020ensemble}, we decompose the GP into local unary factors as a sparse approximation. Given that $\mathbf{y}_i$ and query node $\mathbf{y}_j^*$ follow a GP, the joint distribution and conditional are: 
\begin{equation}
    \label{eq:7}
   \begin{split}
       \begin{bmatrix}\mathbf{y}_i \\ \mathbf{y}_j^*\end{bmatrix} &\sim \mathcal{N}
       \left(\begin{bmatrix}0\\ 0\end{bmatrix}, \begin{bmatrix}k_{ii} + \sigma_{\text{n}}^2I & k_{i*} \\ k_{*i} & k_{**}\end{bmatrix}\right) \\
       \mathbf{y}_j^*\mid\mathbf{y}_i 
       &\sim \mathcal{N} 
       \footnotesize
       \bigg( \underbrace{k_{*i}\left(k_{ii} +  \sigma_{\text{n}}^2I\right)^{-1}\mathbf{y}_i}_{\mu_{\mathbf{y}_j^* \mid \mathbf{y}_i}}, \ 
       \underbrace{k_{**} - k^2_{*i} k^{-1}_{ii}}_{\Sigma_{\mathbf{y}_j^* \mid \mathbf{y}_i}}\bigg)
       \normalsize
   \end{split}
\end{equation}
This gives us a unary Gaussian potential which can be incorporated into a least-square setting: 
\begin{equation}
\mathcal{G}_{ij} = {|| \mathbf{y}_j^* - \mu_{\mathbf{y}_j^* \mid \mathbf{y}_i} ||}^2_{\Sigma_{\mathbf{y}_j^* \mid \mathbf{y}_i}}
\label{eq:8}
\end{equation}
At a timestep $t$, given $n$ measurements $(\mathbf{x}_i, \mathbf{y}_i)$, we add the set of associated factors within a local radius $r$ of each query node's position $\mathbf{x}^*_j$. $r$ represents a tradeoff between speed and reconstruction accuracy, and is empirically set to $15\%$ of the ground-truth object's side length. Thus, for all query nodes, we accumulate a small set of factors:
\begin{equation}
\mathcal{R}_t \triangleq  \bigg\{ \{\mathcal{G}_{ij}\}_{\substack{i = 1 \cdots n \\ j = 1 \cdots \small S^3 \normalsize}} \ \mid \  \lVert \mathbf{x}^*_j -  \mathbf{x}_i \rVert \leq r \bigg\}
\label{eq:9}
\end{equation}
This sparsifies an otherwise intractable optimization, pictorially represented in Figure \ref{fig:gpis} for a 2-D case. Taking the Stanford bunny as an example, we illustrate how a set of noisy surface measurements are converted into local GP factors. The final optimization recovers a posterior SDF mean and uncertainty. More specifically, for the visuo-tactile problem, the \textit{maximum a posteriori} estimation is: %
\begin{figure}[t]
	\centering
	\includegraphics[width=\columnwidth]{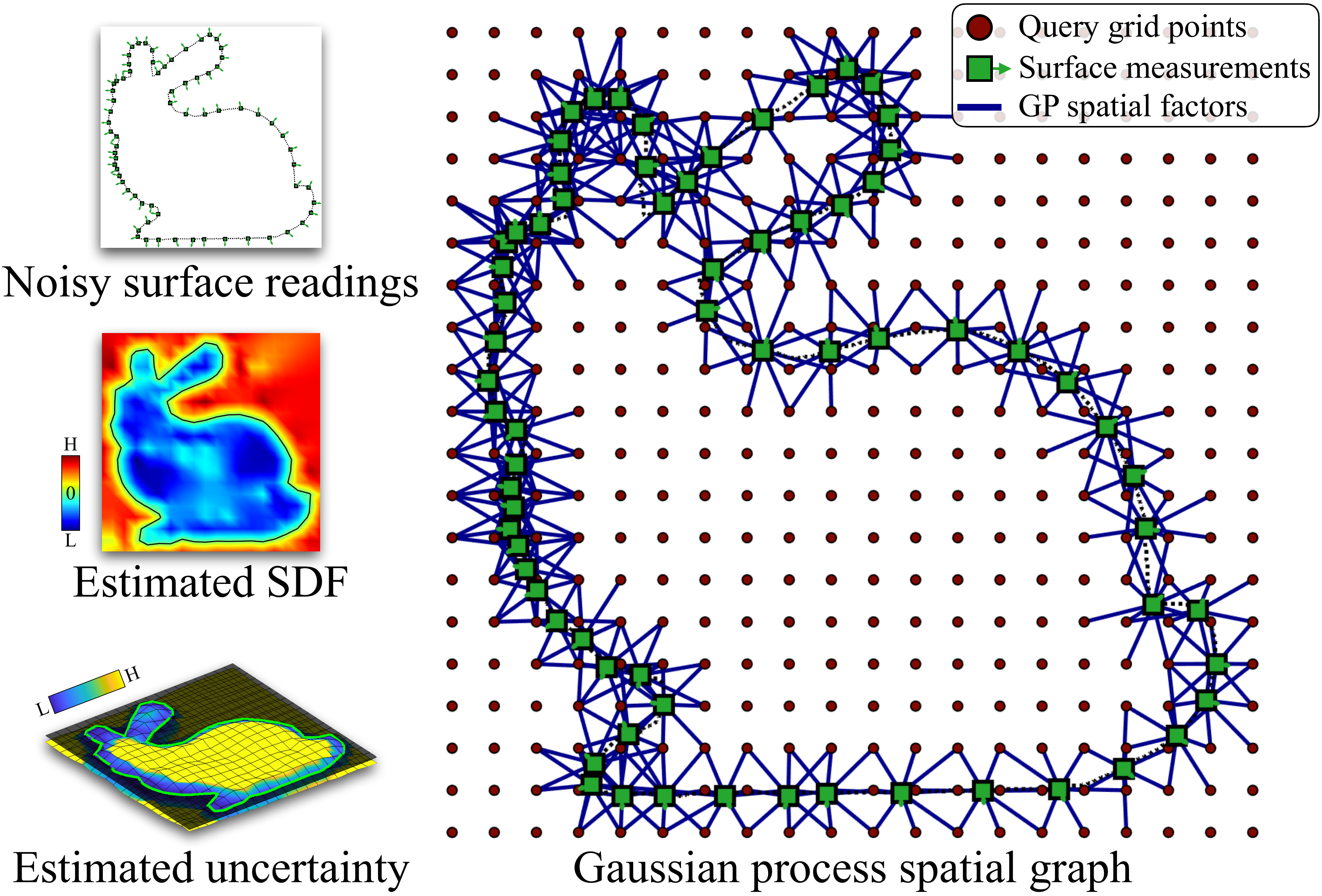}
	\caption{\textbf{[right]} A 2-D illustration of our GP spatial graph (GP-SG), an efficient local approximation to a full GP. The graph consists of SDF query nodes (\protect\tikz[baseline=-0.6ex]\protect\draw[black,fill=query, line width=0.5pt] (0,0) circle (.6ex);) $\mathbf{Y}^*$ each at their spatial positions $\mathbf{X}^*$. Each surface measurement (\protect\tikz[baseline=-.1ex]\protect\draw[black,fill=meas, line width=0.5pt] (0,0) rectangle (1ex,1ex);) ($\mathbf{x}_i, \mathbf{y}_i$) produces a unary factor ({\normalsize \color{factor}{\textbf{--}}}) \footnotesize $\mathcal{G}_{ij}$ at query node $\mathbf{y}_j^*$ (within the local radius $r$). This represents a local Gaussian potential for the GP implicit surface. \textbf{[left]} Optimizing for $\hat{\mathbf{Y}}^*$ yields posterior SDF mean + uncertainty. The zero-level set of the SDF gives us the implicit surface $\mathcal{S}$.}
	\label{fig:gpis} 
	\GobbleLarge
\end{figure}
\begin{figure*}[t]
\centering
\includegraphics[width=0.9\linewidth]{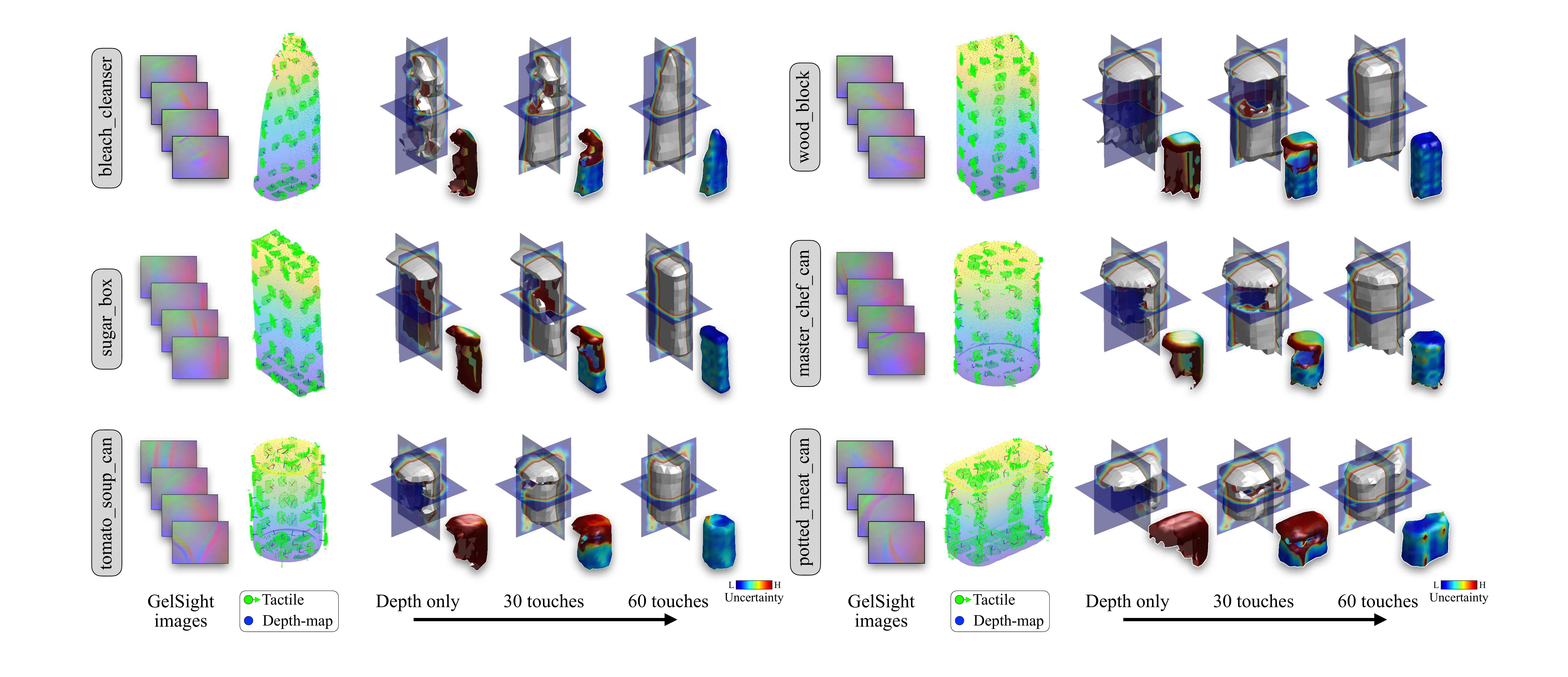}
\caption{Results from simulated visuo-tactile mapping on our \texttt{YCBSight-Sim} dataset. Shown for each object are (i) sample GelSight images, (ii) tactile and depth-map measurements on the ground-truth mesh, and (iii) frames from the incremental mapping. Each object is initialized with a noisy rendered depth-map (\textit{Depth only}), and with each sequential GelSight measurement, we gain further understanding of global shape and reduce surface uncertainty. Visualized here are the implicit surface + SDF + uncertainty for the intervals of $[0, 30, 60]$ touches.}
\label{fig:sim-result} 
\GobbleLarge
\end{figure*}
\begin{equation}
\footnotesize
\begin{split}
\hat{\mathbf{Y}}^* =
\underset{\mathbf{Y^*}}{\operatorname{argmin}}
\overbrace{ \smash[b]{\vphantom{\sum\limits_{i=1}^t }}
\smashoperator{
\sum\limits_{\mathcal{G}_d \in \mathcal{R}_0 }
}
\mathcal{G}_d}^{\text{depth factors}} 
+
\overbrace{
\sum\limits_{t=1}^T  
\smashoperator[r]{
\sum\limits_{\mathcal{G}_m \in \mathcal{R}_t}
}
\mystrut{1.5ex} \mathcal{G}_m
}^{\text{tactile factors}}
+
\overbrace{ \smash[b]{\vphantom{\sum\limits_{i=1}^t }}
\smashoperator[r]{
\sum\limits_{\mathbf{y}^* \in \mathbf{Y^*}}  
}
\mystrut{1.5ex} ||\mathbf{y}^* - \mathbf{b}||_{\Sigma_{\mathbf{b}}}^2}^{\text{GP priors}}
\end{split}
\normalsize
\label{eq:10}
\end{equation}
where $\mathcal{R}_0$ is the factor set from the depth-map $\mathbf{D}_0$, and $\mathcal{R}_t$ is the factor set from tactile measurement $\bm{\mathcal{M}}_t$. The term $\mathbf{b}$ applies a positive SDF prior to nodes, initializing the volume as empty space. Inference is carried out at each timestep via incremental  smoothing  and  mapping  (iSAM2)~\cite{kaess2012isam2}. 

This framework combines the computational benefits of an online local GPIS \cite{lee2019online, stork2020ensemble} with those of an incremental least-squares solver. This is well-suited for sensors like the GelSight, as the dense point-clouds are too expensive to incorporate into a full GP. When querying, we recover the posterior mean and covariance only for the nodes updated---the remaining grid is accessed from cache. 
\subsection{Implicit surface generation} 
\label{ssec:implicit}
The posterior estimate $\hat{\mathbf{Y}}^*$ represents the SDF's mean and uncertainty, sampled from the $S^3$ volume. A marching cubes algorithm \cite{lorensen1987marching} can give us both the implicit surface $\mathcal{S}$ and the corresponding SDF uncertainty $\Sigma_{\mathcal{S}}$. $\mathcal{S}$ is generated as the zero-level set of the SDF: 
\begin{equation}
\mathcal{S} \triangleq \{ \mathbf{s} \in \mathbb{R}^3 \ | \  \mathbf{\hat{Y}^*}(\mathbf{s})_{\varphi} = 0 \}
\label{eq:11}
\end{equation}
Finally, we prune faces/vertices from $\mathcal{S}$ that lie outside $r$ for \textit{any} of the sensor measurements. These areas have high surface uncertainty, and our spatial graph will poorly approximate them. Furthermore, this is necessary for sequential data as we cannot expect a watertight mesh from partial coverage.

\begin{figure}[!b]
	\GobbleLarge
	\centering
	\includegraphics[width=0.9\columnwidth]{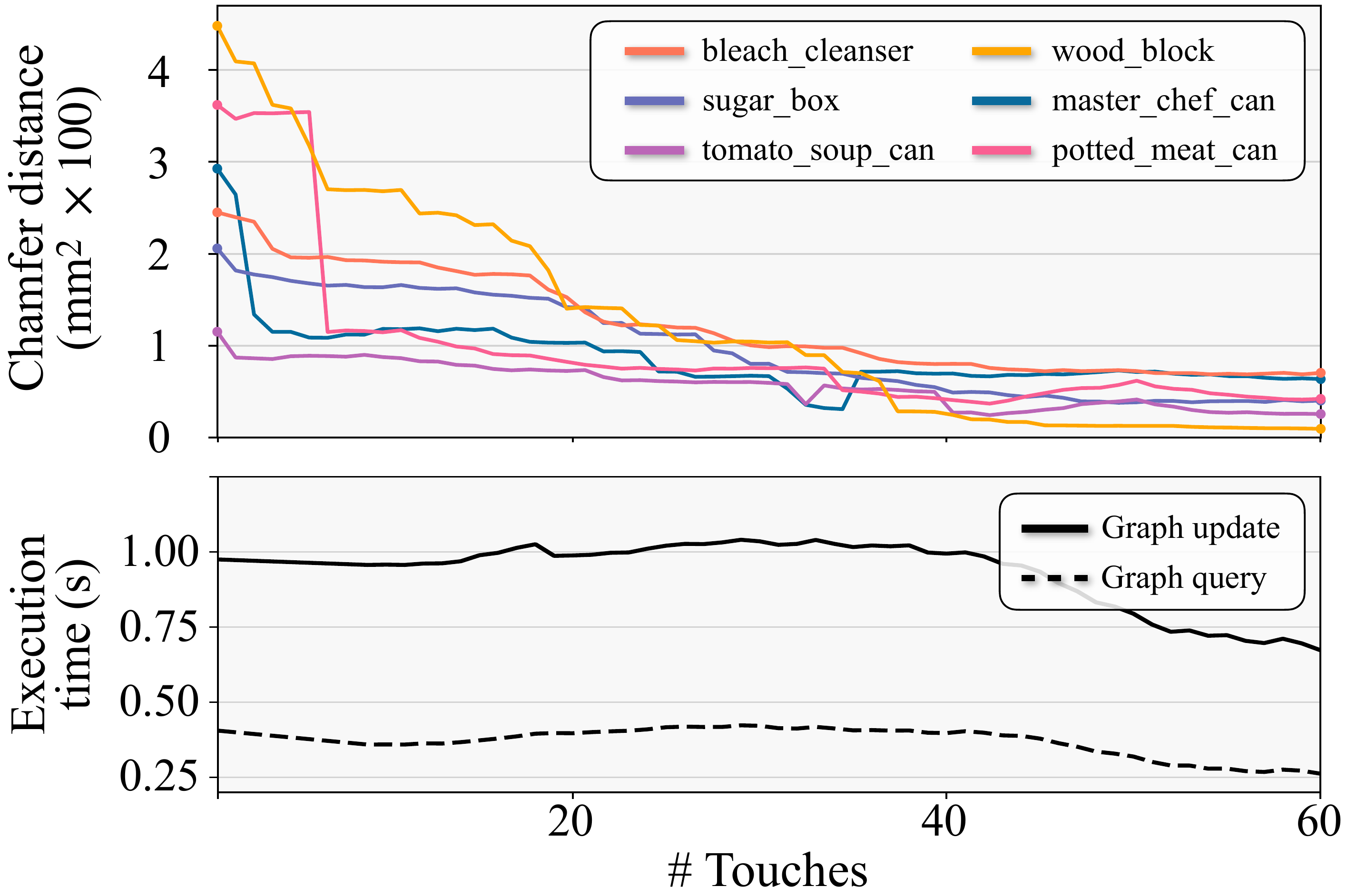}
	\caption{\textbf{[top]} The Chamfer distance (CD) with respect to ground-truth meshes for \texttt{YCBSight-Sim} experiments. Objects are initialized with high CD from partial depth-map, but converge to low-error in 35--40 touches. \textbf{[bottom]} Average execution time for update/query operations on our GP spatial graph (GP-SG). At each touch we add $\approx 10^3$ GP factors during update, and recover posterior mean/covariance during query. We see a dip in timing towards the end, due to smaller contact areas on the top of objects.} 
	\label{fig:sim_graphs} 
	\GobbleMedium
\end{figure}
\begin{figure*}[t]
\centering
\includegraphics[width=0.9\linewidth]{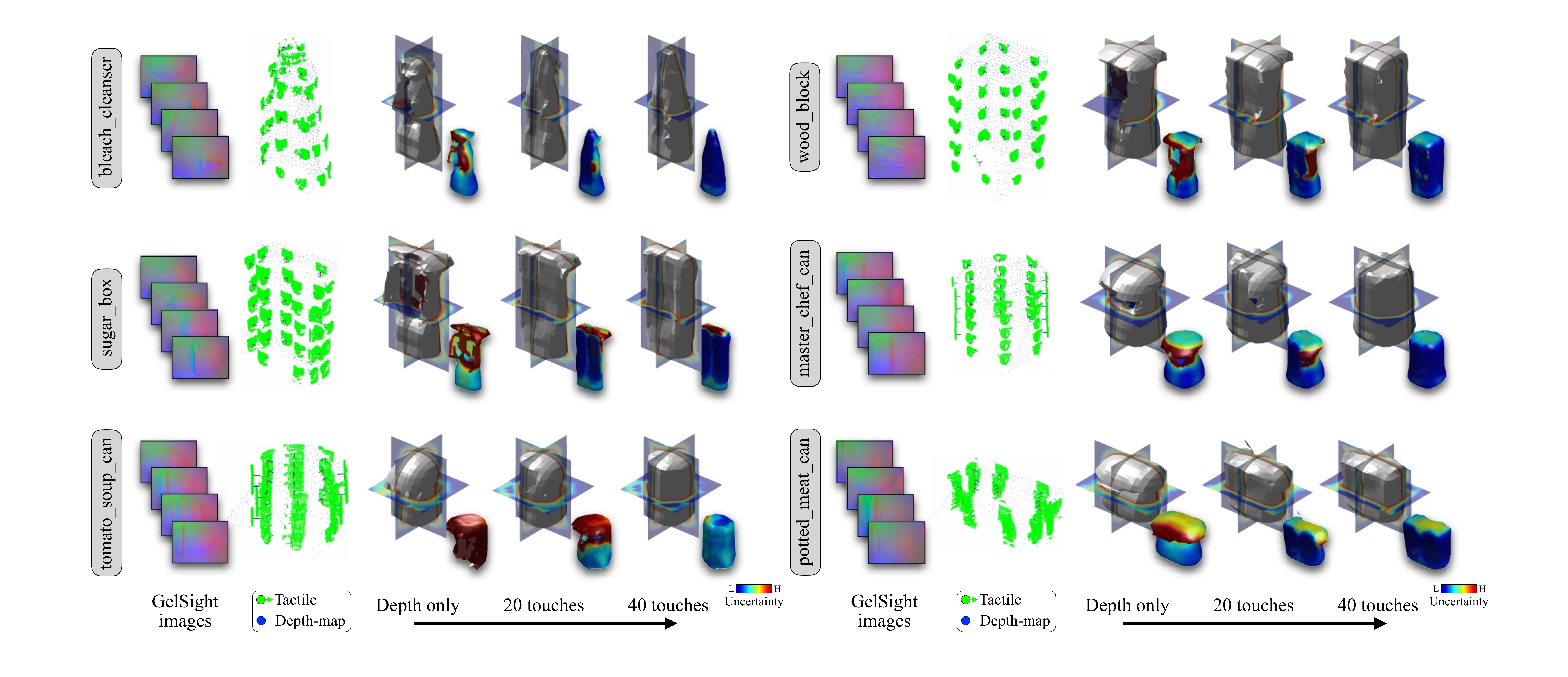}
\caption{Results from real visuo-tactile mapping on our \texttt{YCBSight-Real} dataset. This is structured similar to Figure \ref{fig:sim-result}, except with reconstruction frames at intervals of $[0, 20, 40]$ touches. The Kinect performs poorly for specular objects such as \texttt{tomato\_soup\_can} and \texttt{potted\_meat\_can}, but high-precision GelSight measurements can disambiguate global shape. Our mapping generalizes well and we observe similar results between simulated and real experiments.}
\label{fig:real-result} 
\GobbleMedium
\end{figure*}
\begin{figure}[h]
	\centering
	\includegraphics[width=0.9\columnwidth]{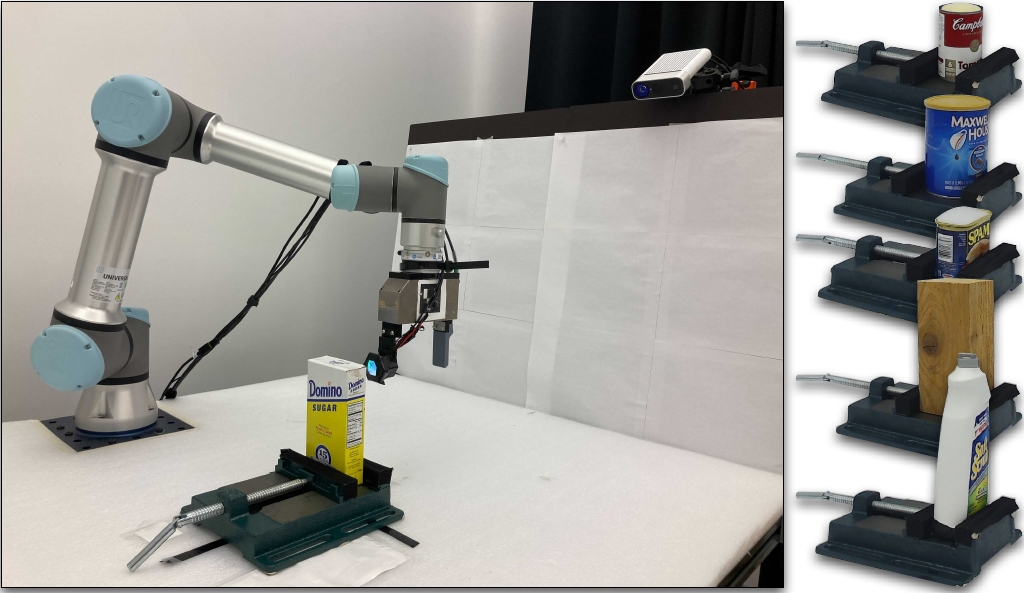}
	\caption{Experimental setup for the \texttt{YCBSight-Real} dataset, with a GelSight tactile sensor, a depth-camera and the YCB objects. Objects are firmly secured on a mechanical bench vise, to ensure they stay stationary. We collect measurements by approaching from a discretized set of angles and heights, and detecting contact from the tactile images. The overlooking Kinect collects a depth-map to initialize our visuo-tactile mapping.}
	\label{fig:real-expt-setup} 
\end{figure}
\begin{figure}[h]
	\GobbleSmall
	\centering
	\includegraphics[width=0.95\columnwidth]{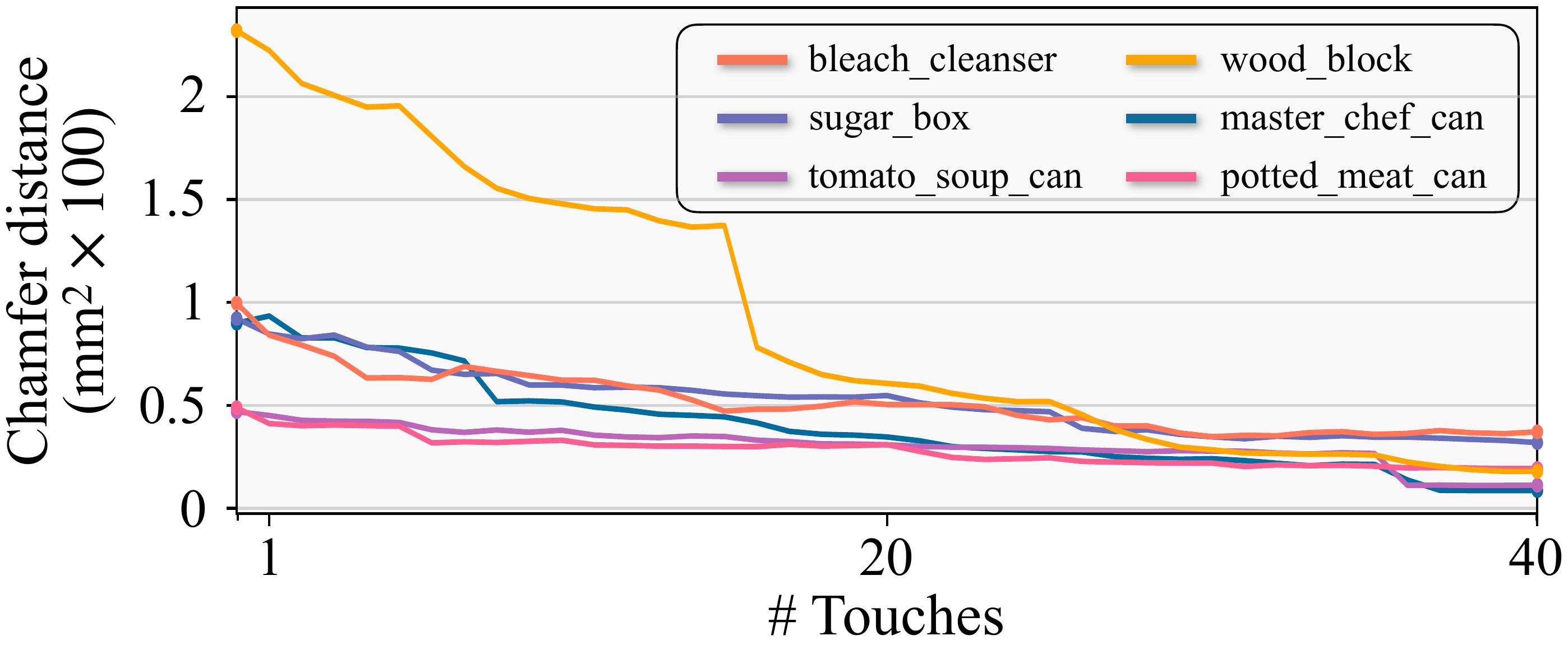}
	\caption{The Chamfer distance (CD) with respect to ground-truth meshes for our \texttt{YCBSight-Real} experiments. We observe the error converges to a similar magnitude as Figure \ref{fig:sim_graphs} after 30 touches. They initially start out with a lower error than simulation as a result of the hallucinated base measurements we add to each object (refer Section \ref{ssec:expts_2}).}
	\label{fig:real_graphs} 
	\GobbleLarge
\end{figure}

\section{Experimental evaluation}
\label{sec:expts}
We illustrate our method in simulated (Section \ref{ssec:expts_1}) and real-world (Section \ref{ssec:expts_2}) visuo-tactile experiments. The shape estimates are compared with respect to the ground-truth using the Chamfer distance (CD) \cite{barrow1977parametric}, a commonly-used shape similarity metric.

\boldsubheading{Implementation}
The framework is executed on an Intel Core i7-7820HQ CPU, 32GB RAM without GPU parallelization. We use the GTSAM \cite{dellaert2012factor} optimizer with iSAM2 \cite{kaess2012isam2} for incremental inference. Due to the precision of sensing, we empirically weight the noise of tactile measurements lower than that of the depth-map. We set the grid size ${\small S = 16}$, which can be increased for higher-resolution reconstructions. 

\subsection{Visuo-tactile data collection}
\label{ssec:dataset}
We collect the \texttt{YCBSight-Sim} and \texttt{YCBSight-Real} datasets for evaluating our method. This comprises of YCB ground-truth meshes \cite{calli2017yale}, GelSight images from interaction, sensor poses, and a depth-map. While we consider 30 household objects in simulation, we restrict our shape mapping evaluation to 6 objects. This subset of objects have varied geometries (curved, rectangular, and complex) to verify the generalization of our method.

\vspace{0.1in}\noindent\texttt{YCBSight-Sim}:
We generate GelSight-object interactions using \textit{Taxim}, an example-based tactile simulator \cite{si2022taxim}. We simulate 60 uniformly spread sensor poses on each object, normal to the local surface of the mesh. We render a depth-map from the perspective of an overlooking camera using Pyrender \cite{pyrender}. Finally, zero-mean Gaussian noise is added to tactile point-clouds, sensor poses, and depth-map. 

\vspace{0.1in}\noindent\texttt{YCBSight-Real}:
We use a UR5e 6-DoF robot arm, mounting the GelSight sensor on a WSG50 parallel gripper. The depth-map is captured via a fixed-pose, calibrated Azure Kinect, approximately $1 \text{m}$ away from the object. Our complete setup can be seen in Figure \ref{fig:real-expt-setup}. The GelSight captures 640~$\times$~480 RGB images of the interactions in a 2.66~cm$^2$ area. The objects are secured by a mechanical bench vise at a known pose, to ensure they remain static. After capturing the depth-map $\mathbf{D}_0$, we approach each object from a discretized set of angles and heights. This simple strategy works well for the selected objects, and future work can replace this with a closed-loop planner. We detect contact events by thresholding the tactile images. We collect $40$ tactile images \small $\{\mathbf{I}_1, {\footnotesize \ldots}, \mathbf{I}_{40}\}$ \normalsize of the object's surface (${\small \approx} \ 20$ minutes) with the corresponding gripper poses \small $\{\mathbf{p}_1,  {\footnotesize \ldots}, \mathbf{p}_{40}\}$ \normalsize via robot kinematics. 

\subsection{Simulated tactile mapping}
\label{ssec:expts_1}

In Figure \ref{fig:sim-result} we highlight mapping results for the 6 objects in \texttt{YCBSight-Sim}. We first visualize the implicit surface and SDF uncertainty from depth-map only. After this, touch measurements are added incrementally and reflect in the shape estimate. The surface uncertainty is typically high for regions that lack depth/tactile information, and reduces over time. Figure \ref{fig:sim_graphs} shows that the CD with respect to the ground-truth mesh decreases with greater number of touches, and converges within 35--40 touches. The timing plot of graph operations shows near-constant graph update and query time. This reduces towards the end of the datasets due to smaller contact areas on the top surface of the objects. These timings can be further improved by parallelizing spatial operations.

\subsection{Real-world tactile mapping}
\label{ssec:expts_2}
In Figure \ref{fig:real-result}, we show our method working on real data collected in \texttt{YCBSight-Real}. The Kinect depth-maps for specular objects like \texttt{tomato\_soup\_can} and \texttt{potted\_meat\_can} are erroneous, but tactile information provides more precise local shape. To prevent damage to the robot and sensor, we do not explore near the base of the object---we instead hallucinate measurements at the bottom based on the nearest corresponding sensor poses. In Figure \ref{fig:real_graphs}, we plot the CD over time for the 6 YCB objects. The initial error is lower than simulation due to the additional hallucinated measurements. We see the error converge to an average CD of 18.3 mm\textsuperscript{2}, a similar magnitude as in the simulated experiments. For reference the average diagonal of the ground-truth YCB objects is $20 \ \text{cm}$.

\section{Conclusion}
\label{sec:conc}
We present an incremental framework for 3-D shape estimation from dense touch and vision. We formulate a GP spatial graph (GP-SG) structure, that efficiently infers an object's implicit surface and SDF uncertainty. To integrate GelSight tactile images, we recover local shape with a model learned in tactile simulation. Our method is first demonstrated in a simulated visuo-tactile setting, and is later shown to generalize to real-world shape perception. 

As future work, we wish to actively reconstruct these shapes using surface uncertainty information. The current method can further benefit from (i) parallelized spatial graph operations, and (ii) data-driven shape priors \cite{varley2017shape, wang20183d}. Finally, we wish to consider relaxing the fixed-pose assumption \cite{Suresh21tactile}, and perception of deformable objects.

\footnotesize
\bibliographystyle{ieeetr}
\bibliography{references}

\begin{thebibliography}{10}

\bibitem{newcombe2011kinectfusion}
R.~A. Newcombe, S.~Izadi, O.~Hilliges, D.~Molyneaux, D.~Kim, A.~J. Davison,
  P.~Kohi, J.~Shotton, S.~Hodges, and A.~Fitzgibbon, ``Kinectfusion: Real-time
  dense surface mapping and tracking,'' in {\em Proc. Int. Symposium on Mixed
  and Augmented Reality (ISMAR)}, pp.~127--136, IEEE, 2011.

\bibitem{helbig2007optimal}
H.~B. Helbig and M.~O. Ernst, ``Optimal integration of shape information from
  vision and touch,'' {\em Experimental brain research}, vol.~179, no.~4,
  pp.~595--606, 2007.

\bibitem{yamaguchi2016combining}
A.~Yamaguchi and C.~G. Atkeson, ``Combining finger vision and optical tactile
  sensing: Reducing and handling errors while cutting vegetables,'' in {\em
  Proc. IEEE-RAS Intl. Conf. on Humanoid Robots (Humanoids)}, pp.~1045--1051,
  IEEE, 2016.

\bibitem{yuan2017gelsight}
W.~Yuan, S.~Dong, and E.~H. Adelson, ``{G}el{S}ight: High-resolution robot
  tactile sensors for estimating geometry and force,'' {\em Sensors}, vol.~17,
  no.~12, p.~2762, 2017.

\bibitem{donlon2018gelslim}
E.~Donlon, S.~Dong, M.~Liu, J.~Li, E.~Adelson, and A.~Rodriguez, ``{G}el{S}lim:
  A high-resolution, compact, robust, and calibrated tactile-sensing finger,''
  in {\em Proc. IEEE/RSJ Intl. Conf. on Intelligent Robots and Systems (IROS)},
  pp.~1927--1934, IEEE, 2018.

\bibitem{ward2018tactip}
B.~Ward-Cherrier, N.~Pestell, L.~Cramphorn, B.~Winstone, M.~E. Giannaccini,
  J.~Rossiter, and N.~F. Lepora, ``The {T}ac{T}ip family: Soft optical tactile
  sensors with {3D}-printed biomimetic morphologies,'' {\em Soft robotics},
  vol.~5, no.~2, pp.~216--227, 2018.

\bibitem{alspach2019soft}
A.~Alspach, K.~Hashimoto, N.~Kuppuswamy, and R.~Tedrake, ``Soft-bubble: A
  highly compliant dense geometry tactile sensor for robot manipulation,'' in
  {\em Proc. IEEE Intl. Conf. on Soft Robotics (RoboSoft)}, pp.~597--604, IEEE,
  2019.

\bibitem{lambeta2020digit}
M.~Lambeta, P.-W. Chou, S.~Tian, B.~Yang, B.~Maloon, V.~R. Most, D.~Stroud,
  R.~Santos, A.~Byagowi, G.~Kammerer, {\em et~al.}, ``{DIGIT}: A novel design
  for a low-cost compact high-resolution tactile sensor with application to
  in-hand manipulation,'' {\em IEEE Robotics and Automation Letters (RA-L)},
  vol.~5, no.~3, pp.~3838--3845, 2020.

\bibitem{padmanabha2020omnitact}
A.~Padmanabha, F.~Ebert, S.~Tian, R.~Calandra, C.~Finn, and S.~Levine,
  ``{O}mni{T}act: A multi-directional high-resolution touch sensor,'' in {\em
  Proc. IEEE Intl. Conf. on Robotics and Automation (ICRA)}, pp.~618--624,
  IEEE, 2020.

\bibitem{wang2021gelsight}
S.~Wang, Y.~She, B.~Romero, and E.~Adelson, ``{G}el{S}ight {W}edge: Measuring
  high-resolution {3D} contact geometry with a compact robot finger,'' in {\em
  Proc. IEEE Intl. Conf. on Robotics and Automation (ICRA)}, IEEE, 2021.

\bibitem{wang20183d}
S.~Wang, J.~Wu, X.~Sun, W.~Yuan, W.~T. Freeman, J.~B. Tenenbaum, and E.~H.
  Adelson, ``{3D} shape perception from monocular vision, touch, and shape
  priors,'' in {\em Proc. IEEE/RSJ Intl. Conf. on Intelligent Robots and
  Systems (IROS)}, pp.~1606--1613, IEEE, 2018.

\bibitem{bauza2019tactile}
M.~Bauza, O.~Canal, and A.~Rodriguez, ``Tactile mapping and localization from
  high-resolution tactile imprints,'' in {\em Proc. IEEE Intl. Conf. on
  Robotics and Automation (ICRA)}, pp.~3811--3817, IEEE, 2019.

\bibitem{smith20203d}
E.~J. Smith, R.~Calandra, A.~Romero, G.~Gkioxari, D.~Meger, J.~Malik, and
  M.~Drozdzal, ``{3D} shape reconstruction from vision and touch,'' in {\em
  Proc. Conf. on Neural Information Processing Systems ({NeurIPS})}, 2020.

\bibitem{smith2021active}
E.~J. Smith, D.~Meger, L.~Pineda, R.~Calandra, J.~Malik, A.~Romero, and
  M.~Drozdzal, ``Active {3D} shape reconstruction from vision and touch,'' in
  {\em Proc. Conf. on Neural Information Processing Systems ({NeurIPS})}, 2021.

\bibitem{hertzmann2005example}
A.~Hertzmann and S.~M. Seitz, ``Example-based photometric stereo: Shape
  reconstruction with general, varying {BRDF}s,'' {\em {IEEE} Trans. Pattern
  Anal. Machine Intell.}, vol.~27, no.~8, pp.~1254--1264, 2005.

\bibitem{retrographic}
M.~K. Johnson and E.~H. Adelson, ``Retrographic sensing for the measurement of
  surface texture and shape,'' in {\em Proc. IEEE Conf. on Computer Vision and
  Pattern Recognition (CVPR)}, pp.~1070--1077, 2009.

\bibitem{microgeometry}
M.~K. Johnson, F.~Cole, A.~Raj, and E.~H. Adelson, ``Microgeometry capture
  using an elastomeric sensor,'' {\em ACM Trans. Graph.}, vol.~30, July 2011.

\bibitem{sodhi2021patchgraph}
P.~Sodhi, M.~Kaess, M.~Mukadam, and S.~Anderson, ``Patchgraph: In-hand tactile
  tracking with learned surface normals,'' in {\em Proc. IEEE Intl. Conf. on
  Robotics and Automation (ICRA)}, 2022.

\bibitem{ambrus2021monocular}
R.~Ambrus, V.~Guizilini, N.~Kuppuswamy, A.~Beaulieu, A.~Gaidon, and A.~Alspach,
  ``Monocular depth estimation for soft visuotactile sensors,'' in {\em Proc.
  IEEE Intl. Conf. on Soft Robotics (RoboSoft)}, 2021.

\bibitem{bauza2020tactile}
M.~Bauza, E.~Valls, B.~Lim, T.~Sechopoulos, and A.~Rodriguez, ``Tactile object
  pose estimation from the first touch with geometric contact rendering,'' in
  {\em Proc. Conf. on Robot Learning, CoRL}, 2020.

\bibitem{agarwal2021simulation}
A.~Agarwal, T.~Man, and W.~Yuan, ``Simulation of vision-based tactile sensors
  using physics based rendering,'' in {\em Proc. IEEE Intl. Conf. on Robotics
  and Automation (ICRA)}, pp.~1--7, IEEE, 2021.

\bibitem{wang2022tacto}
S.~Wang, M.~M. Lambeta, P.-W. Chou, and R.~Calandra, ``{TACTO}: A fast,
  flexible, and open-source simulator for high-resolution vision-based tactile
  sensors,'' {\em IEEE Robotics and Automation Letters (RA-L)}, 2022.

\bibitem{si2022taxim}
Z.~Si and W.~Yuan, ``Taxim: An example-based simulation model for gelsight
  tactile sensors,'' {\em IEEE Robotics and Automation Letters (RA-L)}, 2022.

\bibitem{bjorkman2013enhancing}
M.~Bj{\"o}rkman, Y.~Bekiroglu, V.~H{\"o}gman, and D.~Kragic, ``Enhancing visual
  perception of shape through tactile glances,'' in {\em Proc. IEEE/RSJ Intl.
  Conf. on Intelligent Robots and Systems (IROS)}, pp.~3180--3186, IEEE, 2013.

\bibitem{ilonen2014three}
J.~Ilonen, J.~Bohg, and V.~Kyrki, ``Three-dimensional object reconstruction of
  symmetric objects by fusing visual and tactile sensing,'' {\em Intl. J. of
  Robotics Research (IJRR)}, vol.~33, no.~2, pp.~321--341, 2014.

\bibitem{varley2017shape}
J.~Varley, C.~DeChant, A.~Richardson, J.~Ruales, and P.~Allen, ``Shape
  completion enabled robotic grasping,'' in {\em Proc. IEEE/RSJ Intl. Conf. on
  Intelligent Robots and Systems (IROS)}, pp.~2442--2447, IEEE, 2017.

\bibitem{gandler2020object}
G.~Z. Gandler, C.~H. Ek, M.~Bj{\"o}rkman, R.~Stolkin, and Y.~Bekiroglu,
  ``Object shape estimation and modeling, based on sparse {G}aussian process
  implicit surfaces, combining visual data and tactile exploration,'' {\em J.
  of Robotics and Autonomous Systems (RAS)}, vol.~126, p.~103433, 2020.

\bibitem{williams2007gaussian}
O.~Williams and A.~Fitzgibbon, ``{G}aussian process implicit surfaces,'' {\em
  Gaussian Proc. in Practice}, pp.~1--4, 2007.

\bibitem{bierbaum2008robust}
A.~Bierbaum, I.~Gubarev, and R.~Dillmann, ``Robust shape recovery for sparse
  contact location and normal data from haptic exploration,'' in {\em Proc.
  IEEE/RSJ Intl. Conf. on Intelligent Robots and Systems (IROS)},
  pp.~3200--3205, IEEE, 2008.

\bibitem{dragiev2011gaussian}
S.~Dragiev, M.~Toussaint, and M.~Gienger, ``{G}aussian process implicit
  surfaces for shape estimation and grasping,'' in {\em Proc. IEEE Intl. Conf.
  on Robotics and Automation (ICRA)}, pp.~2845--2850, IEEE, 2011.

\bibitem{ottenhaus2016local}
S.~Ottenhaus, M.~Miller, D.~Schiebener, N.~Vahrenkamp, and T.~Asfour, ``Local
  implicit surface estimation for haptic exploration,'' in {\em Proc. IEEE-RAS
  Intl. Conf. on Humanoid Robots (Humanoids)}, pp.~850--856, IEEE, 2016.

\bibitem{jamali2016active}
N.~Jamali, C.~Ciliberto, L.~Rosasco, and L.~Natale, ``Active perception:
  Building objects' models using tactile exploration,'' in {\em Proc. IEEE-RAS
  Intl. Conf. on Humanoid Robots (Humanoids)}, pp.~179--185, IEEE, 2016.

\bibitem{yi2016active}
Z.~Yi, R.~Calandra, F.~Veiga, H.~van Hoof, T.~Hermans, Y.~Zhang, and J.~Peters,
  ``Active tactile object exploration with {G}aussian processes,'' in {\em
  Proc. IEEE/RSJ Intl. Conf. on Intelligent Robots and Systems (IROS)},
  pp.~4925--4930, IEEE, 2016.

\bibitem{driess2017active}
D.~Driess, P.~Englert, and M.~Toussaint, ``Active learning with query paths for
  tactile object shape exploration,'' in {\em Proc. IEEE/RSJ Intl. Conf. on
  Intelligent Robots and Systems (IROS)}, pp.~65--72, IEEE, 2017.

\bibitem{lee2019online}
B.~Lee, C.~Zhang, Z.~Huang, and D.~D. Lee, ``Online continuous mapping using
  {G}aussian process implicit surfaces,'' in {\em Proc. IEEE Intl. Conf. on
  Robotics and Automation (ICRA)}, pp.~6884--6890, IEEE, 2019.

\bibitem{stork2020ensemble}
J.~A. Stork and T.~Stoyanov, ``Ensemble of sparse {G}aussian process experts
  for implicit surface mapping with streaming data,'' in {\em Proc. IEEE Intl.
  Conf. on Robotics and Automation (ICRA)}, pp.~10758--10764, IEEE, 2020.

\bibitem{ranganathan2010online}
A.~Ranganathan, M.-H. Yang, and J.~Ho, ``Online sparse {G}aussian process
  regression and its applications,'' {\em IEEE Trans. on Image Processing},
  vol.~20, no.~2, pp.~391--404, 2010.

\bibitem{dellaert2017factor}
F.~Dellaert and M.~Kaess, ``Factor graphs for robot perception,'' {\em
  Foundations and Trends in Robotics}, vol.~6, no.~1-2, pp.~1--139, 2017.

\bibitem{yan2017incremental}
X.~Yan, V.~Indelman, and B.~Boots, ``Incremental sparse {GP} regression for
  continuous-time trajectory estimation and mapping,'' {\em J. of Robotics and
  Autonomous Systems (RAS)}, vol.~87, pp.~120--132, 2017.

\bibitem{rosen2014inference}
D.~M. Rosen, G.~Huang, and J.~J. Leonard, ``Inference over heterogeneous
  finite-/infinite-dimensional systems using factor graphs and {G}aussian
  processes,'' in {\em Proc. IEEE Intl. Conf. on Robotics and Automation
  (ICRA)}, pp.~1261--1268, IEEE, 2014.

\bibitem{mukadam2018continuous}
M.~Mukadam, J.~Dong, X.~Yan, F.~Dellaert, and B.~Boots, ``Continuous-time
  {G}aussian process motion planning via probabilistic inference,'' {\em Intl.
  J. of Robotics Research (IJRR)}, vol.~37, no.~11, pp.~1319--1340, 2018.

\bibitem{wang2019underwater}
J.~Wang, T.~Shan, and B.~Englot, ``Underwater terrain reconstruction from
  forward-looking sonar imagery,'' in {\em Proc. IEEE Intl. Conf. on Robotics
  and Automation (ICRA)}, pp.~3471--3477, IEEE, 2019.

\bibitem{Suresh21tactile}
S.~Suresh, M.~Bauza, K.-T. Yu, J.~G. Mangelson, A.~Rodriguez, and M.~Kaess,
  ``Tactile {SLAM}: Real-time inference of shape and pose from planar
  pushing,'' in {\em Proc. IEEE Intl. Conf. on Robotics and Automation (ICRA)},
  May 2021.

\bibitem{eigen2014depth}
D.~Eigen, C.~Puhrsch, and R.~Fergus, ``Depth map prediction from a single image
  using a multi-scale deep network,'' in {\em Advances in Neural Information
  Processing Systems}, vol.~3, pp.~2366--2374, 2014.

\bibitem{laina2016deeper}
I.~Laina, C.~Rupprecht, V.~Belagiannis, F.~Tombari, and N.~Navab, ``Deeper
  depth prediction with fully convolutional residual networks,'' in {\em Proc.
  Intl. Conf. on {3D} Vision (3DV)}, pp.~239--248, IEEE, 2016.

\bibitem{fcrn2018github}
``{FCRN}: Fully convolutional residual network for depth estimation.''
  \texttt{\url{https://github.com/XPFly1989/FCRN}}, 2018.

\bibitem{calli2017yale}
B.~Calli, A.~Singh, J.~Bruce, A.~Walsman, K.~Konolige, S.~Srinivasa, P.~Abbeel,
  and A.~M. Dollar, ``{Y}ale-{CMU}-{B}erkeley dataset for robotic manipulation
  research,'' {\em Intl. J. of Robotics Research (IJRR)}, vol.~36, no.~3,
  pp.~261--268, 2017.

\bibitem{rasmussen2003gaussian}
C.~E. Rasmussen, ``{G}aussian processes in machine learning,'' in {\em Summer
  School on Machine Learning}, pp.~63--71, Springer, 2003.

\bibitem{kaess2012isam2}
M.~Kaess, H.~Johannsson, R.~Roberts, V.~Ila, J.~Leonard, and F.~Dellaert,
  ``i{SAM2}: Incremental smoothing and mapping using the {B}ayes tree,'' {\em
  Intl. J. of Robotics Research (IJRR)}, vol.~31, no.~2, pp.~216--235, 2012.

\bibitem{lorensen1987marching}
W.~E. Lorensen and H.~E. Cline, ``Marching cubes: A high resolution {3D}
  surface construction algorithm,'' {\em {ACM} {SIGGRAPH} Computer Graphics},
  vol.~21, no.~4, pp.~163--169, 1987.

\bibitem{barrow1977parametric}
H.~G. Barrow, J.~M. Tenenbaum, R.~C. Bolles, and H.~C. Wolf, ``Parametric
  correspondence and chamfer matching: Two new techniques for image matching,''
  tech. rep., 1977.

\bibitem{dellaert2012factor}
F.~Dellaert, ``Factor graphs and {GTSAM}: A hands-on introduction,'' tech.
  rep., Georgia Institute of Technology, 2012.

\bibitem{pyrender}
M.~Matl, ``Pyrender.'' \texttt{\url{https://github.com/mmatl/pyrender}}, 2019.

\end{thebibliography}
\end{document}